\documentclass{article} 
\usepackage{iclr2021_conference,times}

\usepackage{hyperref}
\usepackage{url}

\usepackage[utf8]{inputenc} 
\usepackage[T1]{fontenc}              
\usepackage{amsmath}
\usepackage{graphicx}
\usepackage{booktabs}
\usepackage{multirow}
\usepackage{tablefootnote}

\usepackage{mathtools}  
\usepackage{amsthm}     
\usepackage{amssymb}    
\usepackage{mathrsfs}   
\usepackage{amsopn}     
\usepackage{stmaryrd}   
\usepackage{dsfont}     
\usepackage{bm}         
\SetSymbolFont{stmry}{bold}{U}{stmry}{m}{n} 
\usepackage{mycommands}
\usepackage{cleveref}

\DeclareMathOperator*{\Ave}{Ave}

\newcommand\nhalf{-\frac 1 2 }

\newcommand{\sign}{\rm sign}
\newcommand{\rb}{\rangle}
\newcommand{\lb}{\langle}
\newcommand{\OSigma}{\overline \Sigma}

\newcommand{\la}{\lambda}
\newcommand{\R} {{\mathbb R}}

\newcommand{\E} {{\mathbb E}}

\newtheorem{theorem}{Theorem}[section]

\newtheorem{proposition}[theorem]{Proposition}

\newtheorem{lemma}{Lemma}[section]

\title{Separation and Concentration in Deep Networks}

\author{John Zarka,  Florentin Guth\\
  Département d’informatique de l'ENS, ENS, CNRS, PSL University, Paris, France \\
  \texttt{\{john.zarka,florentin.guth\}@ens.fr} \\
\And
St\'ephane Mallat \\
Collège de France, Paris, France\\
Flatiron Institute, New York, USA 
}

\iclrfinalcopy
\begin{document}

\maketitle

\begin{abstract}
Numerical experiments demonstrate that deep neural network classifiers progressively separate class distributions around their mean, achieving linear separability on the training set, and increasing the Fisher discriminant ratio. We explain this mechanism with two types of operators. We prove that a rectifier without biases applied to sign-invariant tight frames can separate class means and increase Fisher ratios. On the opposite, a soft-thresholding on tight frames can reduce within-class variabilities while preserving class means. Variance reduction bounds are proved for Gaussian mixture models. For image classification, we show that separation of class means can be achieved with rectified wavelet tight frames that are not learned. It defines a scattering transform. Learning  $1 \times 1$ convolutional tight frames along scattering channels and applying a soft-thresholding reduces within-class variabilities. The resulting scattering network reaches the classification accuracy of ResNet-18 on CIFAR-10 and ImageNet, with fewer layers and no learned biases.
\end{abstract}

\section{Introduction}
Several numerical works \citep{oyallon, papyan, donoho} have shown that deep 
neural networks classifiers \citep{nature} progressively concentrate each class around separated means, until the last layer, where within-classes variability may nearly ``collapse'' \citep{donoho}. The linear separability of a class mixture is characterized by the Fisher discriminant ratio \citep{fisher, rao}. The Fisher discriminant ratio measures the separation of class means relatively to the variability within
each class, as measured by their covariances. The neural collapse appears through a considerable increase of the Fisher discriminant ratio during training \citep{donoho}.  No mathematical mechanism has yet been provided to explain this separation and concentration of probability measures.

Linear separability and Fisher ratios can be increased by 
separating class means without increasing the variability of each class, or
by concentrating each class around its mean while preserving the mean separation.
This paper shows that these separation or concentration properties can be achieved with one-layer network operators using different pointwise non-linearities. 
We cascade these operators to define structured deep neural networks with high classification accuracies, and which can be analyzed mathematically.

Section \ref{framecontract} studies two-layer networks computed with a
linear classifier applied to $\rho F$, where $F$ is linear and $\rho$ is a pointwise non-linearity.
First, we show that $\rho F$
can separate class means with 
a ReLU $\rho_r (u)=\max(u,0)$ and a sign-invariant $F$. We prove that $\rho_r F$ then 
increases the Fisher ratio.
As in Parseval networks \citep{parseval},  $F$ is normalized by imposing that
it is a tight frame which satisfies $F^T F = \Id$. 
Second, to concentrate the variability of each class around its mean, we 
use a shrinking non-linearity implemented by a soft-thresholding $\rho_t$. 
For Gaussian mixture models, we prove that $\rho_t F$ concentrates within-class variabilities while nearly preserving class means, under appropriate sparsity hypotheses.
A linear classifier applied to these $\rho F$ defines 
two-layer neural networks with no learned bias parameters in the hidden layer,
whose properties are studied mathematically and numerically.

Cascading several convolutional tight frames with ReLUs or soft-thresholdings defines a deep neural network which progressively separates class means and concentrates their variability. One may wonder if we can avoid learning these frames by using prior information on the geometry of images. 
Section \ref{cascade}
shows that the class mean separation can be computed with wavelet tight frames, which are not learned. They separate scales, directions and phases, which are known groups of transformations.
A cascade of wavelet filters and rectifiers defines a scattering transform \citep{mallatscattering}, which has previously been applied to image classification \citep{joan,OyallonICLR}. However, such networks do not reach state-of-the-art classification results. We show that important improvements are obtained
by learning $1 \times 1$ convolutional projectors and tight frames,  
which concentrate within-class variabilities with soft-thresholdings. It
defines a bias-free deep scattering network whose classification accuracy reaches ResNet-18 \citep{resnet} on CIFAR-10 and ImageNet. Code to reproduce all experiments of the paper is available at \url{https://github.com/j-zarka/separation_concentration_deepnets}.

The main contributions of this paper are:
\begin{itemize}
  \item A double mathematical mechanism to separate and concentrate distinct probability measures, with a rectifier and a soft-thresholding applied to tight frames. The increase of Fisher ratio is proved for tight-frame separation with a rectifier. Bounds on within-class covariance reduction are proved for a soft-thresholding on Gaussian mixture models.

  \item The introduction of a bias-free scattering network which reaches ResNet-18 accuracy on CIFAR-10 and ImageNet. Learning is reduced to $1 \times 1$ convolutional tight frames which concentrate variabilities along scattering channels. 
\end{itemize}

\section{Classification by Separation and Concentration} \label{framecontract}
The last hidden layer of a neural network defines a representation $\Phi(x)$, to which is applied a linear
classifier. This section studies
the separation of class means and class variability concentration
for $\Phi = \rho F$ in a two-layer network.

\subsection{Tight Frame Rectification and Thresholding}
\label{Tightsec}

We begin by
briefly reviewing the properties of linear classifiers and Fisher discriminant ratios. We then analyze the separation and concentration
of $\Phi  = \rho F$, when $\rho$ is a rectifier or a soft-thresholding and $F$ is a tight frame.

\paragraph{Linear classification and Fisher ratio}
We consider a random data vector $x \in \R^d$
whose class labels are $y(x) \in \{1,...,C\}$.
Let $x_c$ be a random vector representing the class $c$, whose
probability distribution is the distribution of 
$x$ conditioned by $y(x) = c$.
We suppose that all classes are equiprobable for simplicity.  $\Ave_c$ denotes $C^{-1} \summ cC$.  

We compute a representation of $x$ with an operator $\Phi$ which is standardized, so that
 $\E (\Phi(x)) = 0$ and each coefficient of $\Phi(x)$ has a unit variance.
 The class means $\mu_c = \E(\Phi(x_c))$ thus satisfy
 $\sum_c \mu_c = 0$.
 A linear classifier $(W,b)$ on $\Phi(x)$ returns the index of the maximum coordinate of
$W \Phi(x) + b \in \R^C$. An optimal linear classifier $(W,b)$ minimizes
the probability of a classification error. 
Optimal linear classifiers are estimated
by minimizing a regularized loss function on the training data. Neural networks often use
logistic linear classifiers, which minimize a cross-entropy loss. The standardization of the last layer $\Phi(x)$ is implemented with a batch normalization \citep{batchnorm}.

A linear classifier can have a small error if the typical sets of
each $\Phi(x_c)$ have little overlap, and in particular
if the class means $\mu_c = \E(\Phi(x_c))$ 
are sufficiently separated relatively to the variability of each class.
Under the Gaussian hypothesis, the variability of each class is measured by the
covariance $\Sigma_c$ of $\Phi(x_c)$. 
Let $\Sigma_W = \Ave_c \Sigma_c$ be the average within-class covariance and $\Sigma_B = \Ave_c \mu_c \, \mu_{c}^T$ be the between-class covariance of the means.
The within-class
covariance can be whitened and normalized to $\Id$ by transforming $\Phi(x)$
with the square root $\Sigma_W^{\nhalf}$ of $\Sigma_W^{-1}$. 
All classes $c,c'$ are highly separated if $\|\Sigma_W^{\nhalf} \mu_{c} - \Sigma_W^{\nhalf} \mu_{c'}\| \gg 1$. This separation is captured by the Fisher discriminant
ratio $\Sigma_W^{-1} \Sigma_B$. We shall measure its trace:
\begin{equation}
\label{Ratio}
C^{-1}\, \Tr (\Sigma_W^{-1} \Sigma_B) = \Ave_c \|\Sigma_W^{\nhalf} \mu_c  \|^2 .
\end{equation}
Fisher ratios have been used to train deep neural networks as a replacement for the cross-entropy loss \citep{deeplda, gerda, fpcanet, deeplda-personid, fisher-melanoma, fihsher-fabricdefect}. In this paper, we use their analytic expression
to analyze the improvement of linear classifiers. 

Linear classification obviously cannot be improved with a linear representation $\Phi$. The following proposition gives a simple condition to improve (or maintain) the
error of linear classifiers with a non-linear representation. 

\begin{proposition}
\label{prop1}
 If $\Phi$ has a linear inverse, then it decreases (or maintains) the error of
 the optimal linear classifier, and it increases (or maintains) the Fisher ratio (\ref{Ratio}).
\end{proposition}

To prove this result, observe that if $\Phi$ has a linear inverse $\Phi^{-1}$ then $W x = W' \Phi(x)$ with
$W' = W \Phi^{-1}$. The minimum classification error by optimizing $W$ is 
thus above the error obtained by optimizing $W'$. \Cref{prop_proof} proves 
that the Fisher ratio (\ref{Ratio}) is also increased or preserved.

There are qualitatively two types of non-linear operators that increase the 
Fisher ratio $\Sigma_W^{-1} \Sigma_B$. Separation operators typically increase the distance between the class means without increasing the variance $\Sigma_W$  within each class. We first study such operators having a linear inverse, which guarantees through Proposition \ref{prop1} that they increase the Fisher ratio. We then study concentration operators which reduce the variability $\Sigma_W$ with non-linear shrinking operators, which are not invertible. It will thus require a finer analysis of their properties.

\paragraph{Separation by tight frame rectification}
Let $\Phi = \rho F$ be an operator which computes the first layer of a neural network, where $\rho$ is a pointwise non-linearity and $F$ is linear.
We first study 
separation operators computed with a ReLU $\rho_r (u) = \max(u,0)$
applied to an invertible sign-invariant matrix. Such a matrix has rows that
can be regrouped in pairs of opposite signs. It can thus be written $F = [-\tilde F^T , \tilde F^T]^T$ where $\tilde F$ is invertible. The operator $\rho F$ separates coefficients according to their sign.
Since $\rho_r (u) - \rho_r (-u) = u$, it results that $\Phi = \rho_r F$
is linearly invertible. According to Proposition \ref{prop1}, it increases (or maintains) the Fisher ratio, and we want to choose $F$ to maximize this increase.

Observe that $\rho_r (\alpha u) = \alpha \rho_r (u)$ if $\alpha \geq 0$. We can thus normalize the rows $f_m$ of $F$ without affecting linear classification performance. To ensure that $F \in \R^{p \times d}$ is invertible with a stable inverse, 
we impose that it is a normalized tight frame of $\R^d$ satisfying
\[
F^T  F = \Id~~\mbox{and}~~\|f_m \|^2 = d/p~\mbox{for $1 \leq m \leq p$}.
\]
 The tight frame can be interpreted as a rotation operator in a higher dimensional space, which aligns the axes and the directions along which $\rho_r$ performs the sign separation. This rotation must be adapted in order to optimize the separation of class means. The fact that $F$ is a tight frame can be interpreted as a normalization which  simplifies the mathematical analysis.
 
  Suppose that all classes $x_c$ of $x$ have a Gaussian distribution with a
 zero  mean $\mu_c = 0$, but different covariances $\Sigma_c$. These classes
 are not linearly separable because they have the same mean, and the Fisher ratio is $0$. Applying $\rho_r F$ can separate these classes and improve the Fisher ratio. Indeed, if $z$ is a zero-mean Gaussian random variable, then  $\E(\max(z,0)) = (2 \pi)^{-1/2} \E(z^2)^{1/2}$ so we verify that for
 $F = [-\tilde F^T , \tilde F^T]^T$,
\[
\E( \rho_r F x_c) = (2 \pi)^{-1/2} \Big( {\rm diag} (\tilde F \Sigma_c \tilde F^T)^{1/2} , {\rm diag} (\tilde F \Sigma_c \tilde F^T)^{1/2} \Big) . 
\]
The Fisher ratio can then be optimized by
maximizing the covariance $\Sigma_B$ between the mean vector components ${\rm diag} (\tilde F \Sigma_c F^T)^{1/2}$ for all classes $c$. If we know a priori that that $x_c$ and $-x_c$ have the same probability distribution, as in the Gaussian example,
then we can replace $\rho_r$ by the
absolute value $\rho_a(u) = |u| = \rho_r(u) + \rho_r(-u)$, 
and $\rho_r F$ by $\rho_a \tilde F$, which reduces by $2$ the frame size.

\paragraph{Concentration by tight frame soft-thresholding} 
If the class means of $x$ are already separated, 
then we can increase the Fisher ratio with a non-linear $\Phi$ that concentrates each class around its mean. The operator $\Phi$ must reduce the within-class variance while preserving the class separation. This can be interpreted as a non-linear noise removal
if we consider the within-class variability 
as an additive noise relatively to the class mean. It can be done with soft-thresholding estimators introduced in \citet{donohojohnstone}.
A soft-thresholding 
$\rho_t (u) = \sign(u)\,\max(|u|-\lambda , 0)$ shrinks the amplitude of $u$ by $\lambda$ in order to reduce its variance, while introducing a bias that depends on $\lambda$.
\citet{donohojohnstone} proved that
soft-thresholding estimators are highly effective to estimate signals that
have a sparse representation in a tight frame $F$.

To evaluate more easily 
the effect of a tight frame soft-thresholding on the class means, we apply 
the linear reconstruction $F^T$ on $\rho_t F x$, which thus defines a representation $\Phi(x) = F^T \rho_t F x$. For a strictly positive threshold, this operator is not invertible, so we cannot apply Proposition \ref{prop1} to prove that the Fisher ratio increases. We  study directly the impact of $\Phi$ on the mean and covariance of each class. 
Let $x_c$ be the vector representing the class $c$. The mean
$\mu_c = \E(x_c)$ is transformed into $\bar \mu_c = \E(\Phi(x_c))$ and
the covariance $\Sigma_c$ of $x_c$ into the covariance $\OSigma_c$ of $\Phi(x_c)$. The average covariances are $\Sigma_W = \Ave_c \Sigma_c$
and $\OSigma_W = \Ave_c \OSigma_c$.

Suppose that
each $x_c$ is a Gaussian mixture, with
a potentially large number of Gaussian components centered at $\mu_{c,k}$ 
with a fixed covariance $\sigma^2 \Id$: 
\begin{equation}
\label{mixture}
p_c = \sum_k \pi_{c,k} \, {\cal N}(\mu_{c,k} , \sigma^2 \Id) .
\end{equation}
This model is quite general, since it amounts to covering the typical set of realizations of $x_c$ with a union of balls of radius $\sigma$, centered in the $(\mu_{c,k})_k$. The following theorem 
relates the reduction of within-class covariance to the sparsity of
$F \mu_{c,k}$. It relies on the soft-thresholding estimation results of \citet{donohojohnstone}. 

For simplicity, we suppose that the tight frame is an orthogonal basis, but the result can be extended to general normalized tight frames. The sparsity is expressed through the decay of
sorted basis coefficients. For a vector $z \in \RR^d$, we denote $z^{(r)}$ a coefficient of rank $r$: 
$|z^{(r)}| \geq |z^{(r+1)}|$
for $1 \leq r \leq d$. The theorem imposes a condition on the amplitude decay of the $(F\mu_{c,k})^{(r)}$ when $r$ increases, which is a sparsity measure.
We write $a(r) \sim b(r)$ if 
$C_1\, a(r) \leq b(r) \leq C_2\, a(r)$ where 
$C_1$ and $C_2$ do not depend upon $d$ nor $\sigma$.
The theorem derives upper bounds on
the reduction of within-class covariances and on the
displacements of class means. The constants
do not depend upon $d$ when it increases to $\infty$ nor on $\sigma$ when it decreases to $0$.

\begin{theorem}
\label{th1}
Under the mixture model hypothesis (\ref{mixture}), we have:
\begin{equation}
\label{tracW}
\Tr(\Sigma_W) = \Tr(\Sigma_M) + \sigma^2\, d ,~~\mbox{with}~~
\Tr(\Sigma_M) = C^{-1}\sum_{c,k} \pi_{c,k}\,\|\mu_c - \mu_{c,k}\|^2.
\end{equation}
If there exists $s > 1/2$ such that 
$|(F\mu_{c,k})^{(r)}| \sim r^{-s}$ then a tight frame soft-thresholding with threshold
$\la = \sigma\, \sqrt{2\log d}$ satisfies:
\begin{equation}
\label{tracWbar}
\Tr( \OSigma_W) = 2 \Tr(\Sigma_M) + O(\sigma^{2-1/s}\, \log d) ,
\end{equation}
and all class means satisfy:
\begin{equation}
\label{meanbar}
\|\mu_c - \overline \mu_c \|^2 = O(\sigma^{2-1/s}\, \log d) .
\end{equation}
\end{theorem}

Under appropriate sparsity hypotheses,
the theorem proves that applying
$\Phi = F^T \rho_t F$ reduces considerably the trace of the within-class covariance. The Gaussian variance $\sigma^2 d$ is dominant
in (\ref{tracW}) and is reduced to $O(\sigma^{2-1/s}\,\log d)$ in 
(\ref{tracWbar}).
The upper bound (\ref{meanbar}) also proves
that $ F^T \rho_t F$ creates a relatively small
displacement of class means, which is proportional to $\log d$. 
This is important to ensure that all class means remain well separated. These
bounds qualitatively explains the increase of Fisher ratios, but they are not sufficient to prove a precise bound on these ratios. 

In numerical experiments, the threshold value of the theorem is automatically adjusted as follows.
Non-asymptotic optimal threshold values have been tabulated as a function of $d$ by \citet{donohojohnstone}. For the
range of $d$ used in our applications, a nearly optimal
threshold is $\lambda = 1.5 \,\sigma$. We rescale 
the frame variance $\sigma^2$ by standardizing
the input $x$ so that it has
a zero mean and each coefficient has a unit variance. 
In high dimension $d$, the within-class variance typically dominates the variance between class means. Under the unit variance assumption we have
$\Tr(\Sigma_W) \approx d$. If $F \in \R^{p \times d}$ is a normalized
tight frame then we also verify as in (\ref{tracW}) that
$\Tr(\Sigma_W) \approx \sigma^2 p$ so $\sigma^2 \approx d/p$. It results
that we choose  $\lambda = 1.5\, \sqrt{d/p}$. 

A soft-thresholding can also be computed from a
ReLU with threshold $\rho_{rt} (u) = \max(u-\lambda,0)$ because
$\rho_t (u) = \rho_{rt} (u) - \rho_{rt} (-u)$. It results that
$[F^T , -F^T] \,\rho_{rt}\, [F^T,-F^T]^T = F^T \,\rho_t\, F$. 
However, a thresholded rectifier has more flexibility than a soft-thresholding, because it may recombine differently $\rho_{rt} F$ and $\rho_{rt} (-F)$ to also separate class means, as explained previously. 
The choice of threshold then becomes a trade-off 
between separation of class means and concentration of class variability.
In numerical experiments, we choose a lower
$\lambda = \sqrt{d/p}$ for a ReLU with a threshold.

\subsection{Two-Layer Networks without Bias}
\label{two-layers}

We study two-layer bias-free networks that implement a linear classification on $\rho F$, where $F$ is a normalized tight frame and $\rho$ may be a rectifier, an absolute value or a soft-thresholding, with no learned bias parameter. Bias-free networks have been introduced for denoising in \citet{biasfreedenoising}, as opposed to classification or regression. We show that such bias-free networks have a limited expressivity and do \emph{not} satisfy universal approximation theorems \citep{apinkus, bach}. However, numerical results indicate that their separation and contractions capabilities are sufficient to reach similar classification results as two-layer networks with biases on standard image datasets.

Applying a linear classifier on $\Phi(x)$ computes:
\[
W \Phi (x) + b = W 
\rho F x + b .
\]
This two-layer neural network has no learned bias parameters in the hidden layer, and we impose that $F^T F = \Id$ with frame rows $(f_m)_m$ having constant norms.
As a result, 
the following theorem proves that it does not satisfy the universal approximation theorem. We define a binary classification problem for which the probability of error  remains above $1/4$ for any number $p$ of neurons in the hidden layer. The proof is provided in \Cref{proof} for a ReLU $\rho_{rt}$ with any threshold. The theorem remains valid with an absolute value $\rho_a$ or a soft-thresholding $\rho_t$, because they are linear combinations of $\rho_{rt}$. 

\begin{theorem}
\label{counter}
Let $\lambda \geq 0$ be a fixed threshold and $\rho_{rt} (u) = \max(u-\lambda,0)$.
Let $\cal F$ be the set of matrices $F \in \R^{p \times d}$ with bounded rows $\|f_m \| \leq 1$.
There exists a random vector $x \in \RR^d$ which admits a probability density supported on the unit ball, and a $C^\infty$ function $h\colon \RR^d \rightarrow \RR$ such that, for all $p \geq d$:
$$\inf_{W \in \R^{1 \times p}, F \in {\cal F}, b \in \R} \mathbb P\bracket{\sgn\paren{W \rho_{rt} F x + b} \neq \sgn(h(x))} \geq \frac14 \, .$$
  \label{th:no-universal-approx}
\end{theorem}

\paragraph{Optimization}
The parameters $W$, $F$ and $b$ are optimized with a stochastic gradient descent that minimizes a logistic cross-entropy loss on the output.
To impose $F^T F = \Id$, following the optimization of Parseval networks \citep{parseval}, after each gradient update of all network parameters, we insert a second gradient step to minimize $\alpha/2 \, \|F^T F - \Id\|^2 $. This gradient update is:
\begin{equation}
  \label{Parseval}
F \leftarrow (1 + \alpha) F - \alpha F F^T F .
\end{equation}
We also make sure after every Parseval step that each tight frame row $f_m$ keeps a constant norm $\norm{f_m}=\sqrt{d/p}$ by applying a spherical projection: $f_m \leftarrow\sqrt{d/p}\, f_m/\norm{f_m}$.
These steps are performed across all experiments described in the paper, which ensures that all singular values of every learned tight frame are comprised between 0.99 and 1.01.

To reduce the number of parameters of the classification matrix $W \in \R^{C \times p}$, we
factorize $W = W'\,F^T$ with $W' \in \R^{C \times d}$. It amounts to reprojecting $\rho F$ in $\R^d$ with the semi-orthogonal frame synthesis $F^T$, and thus defines:
\[
\Phi(x) = F^T\, \rho\, F x .
\]
A batch normalization is introduced after $\Phi$ to stabilize the learning of $W'$.

\begin{figure}[ht]
    \centering
    \caption{Examples of filters $f_m$ from the convolutional tight frame $F$ learned directly on the input $x$ for CIFAR-10, using an absolute value non-linearity $\rho_a$. They resemble wavelet filters.}
    \includegraphics[width=.7\textwidth]{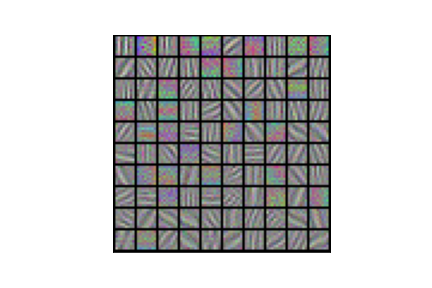}
    \label{fig:atoms}
\end{figure}

\paragraph{Image classification by separation and concentration}
Image classification is first evaluated on the MNIST \citep{mnist} and CIFAR-10 \citep{cifar} image datasets. Table \ref{results} gives the results of logistic classifiers applied to the input signal $x$ and to $\Phi(x) = F^T \rho F x$ for 3 different non-linearities $\rho$: absolute value $\rho_a$, soft-thresholding $\rho_t$, and ReLU with threshold $\rho_{rt}$. The tight frame $F$ is a convolution on patches of size $k \times k$ with a stride of $k/2$, with $k = 14$ for MNIST and $k = 8$ for CIFAR. The tight frame $F$ maps each patch to a vector of larger dimension, specified in \Cref{expdetails}. \Cref{fig:atoms} in \Cref{expdetails} shows examples of learned tight frame filters.

On each dataset, applying $F^T \rho F$ on $x$ greatly reduces linear classification error, which also appears with an increase of the Fisher ratio.
For MNIST, all non-linearities produce nearly the same classification accuracy, but on CIFAR, the soft-thresholding has a higher error. Indeed, the class means of MNIST are distinct averaged digits, which are well separated, because all digits are centered in the image. Concentrating variability with a soft-thresholding is then sufficient. On the opposite, the classes of CIFAR images define nearly stationary random vectors because of arbitrary translations. As a consequence, the class means $\mu_c$ are
nearly constant images, which are only discriminated by their average color. Separating these class means is then important for improving classification. As explained in Section \ref{Tightsec}, this is done by a ReLU $\rho_r$, or in this case an absolute value $\rho_a$, which reduces the error. The ReLU with threshold $\rho_{rt}$ can interpolate between mean separation and variability concentration, and thus performs usually at least as well as the other non-linearities.

The error of the bias-free networks with a ReLU and an absolute value are similar to the
errors obtained by training two-layer networks of similar sizes but with bias parameters: 1.6\% error on MNIST \citep{mnistbench}, and 25\% on CIFAR-10 \citep{cifarbench}. It indicates that the elimination of bias parameters does not affect performances, despite the existence of the counter-examples from Theorem \ref{counter} that cannot be well approximated by such architectures. This means that image classification
problems have more structure that are not captured by these counter-examples, and
that completeness in linear high-dimensional functional spaces may not be key mathematical properties to explain the preformances of neural networks.
\Cref{fig:atoms} in \Cref{expdetails} shows that the learned convolutional tight frames include oriented oscillatory filters, which is also often the case of the first layer of deeper networks \citep{alexnet}. They resemble wavelet frames, which are studied in the next section.

\begin{table}[t]
\caption{For MNIST and CIFAR-10, the first row gives the logistic classification error and the second row the Fisher ratio (\ref{Ratio}), for different signal representations $\Phi(x)$. Results are evaluated with an absolute value $\rho_a$, a soft-thresholding $\rho_t$, and a ReLU with threshold $\rho_{rt}$.}
\label{results}
\bigskip
\makebox[\textwidth][c]{
\begin{tabular}{l l  c c c c c c}
\toprule
\multirow{2}{*}{} & \multirow{2}{*}{$\Phi(x)$} & \multirow{2}{*}{$x$} & \multicolumn{3}{c}{$F^T \rho F x$} & \multirow{2}{*}{$S_T(x)$} \\
\cmidrule(lr){4-6} & & & $\rho = \rho_a$ & $\rho = \rho_t$ & $\rho = \rho_{rt}$  \\
\midrule
\multirow{2}{*}{\bf MNIST} & Error ($\%$) &   7.4 &        1.3 &     1.4 & 1.3 & 0.8 \\
                           & Fisher       &    19 &         68 &      69 &  67 & 130 \\
\midrule
\multirow{2}{*}{\bf CIFAR} & Error ($\%$) &  60.5 &       28.1 & 34.8 & 26.5 & 27.7 \\
                               & Fisher   &   6.7 &         15 &   13 &   16 & 12 \\
\bottomrule
\end{tabular}
}
\end{table}

\section{Deep Learning by Scattering and Concentrating}
\label{cascade}

To improve classification accuracy, we cascade mean separation and variability concentration operators, implemented by ReLUs and soft-thresholdings on tight frames. This defines deep convolutional networks. However, we show that some tight frames do not need to be learned. Section \ref{scat-sec} reviews scattering trees, which perform mean separation by cascading ReLUs on wavelet tight frames. Section \ref{singl-learn} shows that we reach high classification accuracies by learning projectors and tight frame soft-thresholdings, which concentrate within-class variabilities along scattering channels.

\subsection{Scattering Cascade of Wavelet Frame Separations}
\label{scat-sec}
 
Scattering transforms have been introduced to classify images by cascading predefined wavelet filters with a modulus or a rectifier non-linearity
\citep{joan}. We write it as a product of wavelet tight frame rectifications, which progressively separate class means. 

\paragraph{Wavelet frame} A wavelet frame separates image variations at different scales, directions and phases, with a cascade of filterings and subsamplings. We use steerable wavelets \citep{simoncelli1995steerable} computed with Morlet filters \citep{joan}. There is one low-pass filter $g_0$, and $L$ complex band-pass filters $g_\ell$ having an angular direction $\theta = \ell \pi / L$ for $0 < \ell \leq L$.
These filters can be adjusted \citep{dualtreewavelets} so that the filtering and subsampling:
\[
\tilde F_w x(n,\ell) = x \star g_\ell (2 n)
\]
defines a complex tight frame $\tilde F_w$.
Fast multiscale wavelet transforms are computed by cascading the filter bank $\tilde F_w$ on the output of the low-pass filter $g_0$ \citep{mallatbook}. 

Each complex filter $g_\ell$ is analytic, and thus has a real part and imaginary part whose phases are shifted by $\alpha = \pi/2$. This property is important to preserve equivariance to translation despite the subsampling with a stride of $2$
\citep{dualtreewavelets}. To define a sign-invariant frame as in Section \ref{Tightsec}, we must incorporate
filters of opposite signs, which amounts to shifting their phase by $\pi$. 
We thus associate to $\tilde F_w$ a real sign-invariant
tight frame $F_w$ by considering separately the four phases $\alpha = 0,\pi/2,\pi,3\pi/2$. It is defined by
\[
F_w x (n,\ell,\alpha) = x \star g_{\ell,\alpha} (2n) ,
\]
with $g_{\ell,0} = 2^{-1/2} {\rm Real}(g_\ell)$, $g_{\ell,\pi/2} = 2^{-1/2} {\rm Imag}(g_\ell)$ and $g_{\ell,\alpha+\pi} = -g_\ell$. 
We apply a rectifier $\rho_r$ to the output of all real band-pass filters $g_{\ell,\alpha}$
but not to the low-pass filter:
\[
\rho_r F_w  = \Big(  x \star g_{0}(2n)~,~\rho_r (x \star g_{\ell,\alpha}(2n)) \Big)_{n,\alpha,\ell} .
\]
The use of wavelet phase parameters with rectifiers is studied in \citet{zhangRochette}.
The operator $\rho_r F_w$ is linearly invertible because $F_w$ is a tight frame and the ReLU is applied to band-pass filters, which come in pairs of opposite sign. 
Since there are $4$ phases and a subsampling with a stride of 2, $F_w x$ is $(L + 1/4)$ times larger than $x$. 

\paragraph{Scattering tree}
A full scattering tree $S_T$ of depth $J$ is computed by iterating $J$ times over $\rho_r F_w$. Since each $\rho_r F_w$ has a linear inverse, Proposition \ref{prop1} proves that
this separation can only increase the Fisher ratio. However
 it also increases the signal size by $(L + 1/4)^J$, which is typically much too large. This is avoided with orthogonal projectors, which perform a dimension reduction after applying each $\rho_r F_w$. 
 
A pruned scattering tree $S_T$ of depth $J$ and order $o$ is defined in \citet{joan} as a convolutional tree which cascades $J$ rectified wavelet filter banks, and at each depth prunes the branches with $P_j$ to prevent an exponential growth:
\begin{equation}
    \label{scatree}
 S_T = \prod_{j=1}^J P_j\, \rho_r\, F_w .
\end{equation}
After the ReLU, the pruning operator $P_j$ eliminates the branches of the scattering which cascade more than $o$ band-pass filters and rectifiers, where $o$ is the scattering order \citep{joan}. After $J$ cascades, the remaining channels have thus been filtered by at least $J - o$ successive low-pass filters $g_0$. We shall use a scattering transform of order $o = 2$. The operator $P_j$ also averages 
the rectified output
of the filters $g_{\ell,\alpha}$ along the phase $\alpha$, for $\ell$ fixed.
This averaging eliminates the phase. It approximatively computes a complex modulus and
produces a localized translation invariance. The resulting pruning and phase
average operator $P_j$ is a $1 \times 1$ convolutional operator, which reduces the dimension
of scattering channels with an orthogonal projection.
If $x$ has $d$ pixels, then $S_T(x)[n,k]$ is an array of images having $2^{-2J} d$ pixels at each channel $k$, because of the $J$ subsamplings with a stride of $2$.
The total number of channels $K$ is $1 + J L + J (J-1) L^2 / 2$.
Numerical experiments are performed with wavelet filters which approximate Gabor wavelets \citep{joan}, with $L = 8$ directions. The number of scales $J$ depends upon the image size. It is $J = 3$ for MNIST and CIFAR, and $J = 4$ for ImageNet, resulting in respectively $K=217$, $651$ and $1251$ channels.

Each $\rho_r F_w$ can only improve the Fisher ratio and the linear classification accuracy, but it is not guaranteed that this remains valid after applying
$P_j$. 
Table \ref{results} gives the classification error of a logistic classifier applied on $S_T(x)$, after a $1 \times 1$ orthogonal projection to reduce the number of channels, and a spatial normalization. This error is almost twice smaller than a two-layer neural network on MNIST, given in Table \ref{results}, but it does not improve the error on CIFAR. On CIFAR, the error obtained by a ResNet-20 is $3$ times lower than the one of a classifier on $S_T(x)$.
The main issue is now to understand where this inefficiency comes  from.
 
\begin{table}[t]
\caption{Linear classification error and Fisher ratios (\ref{Ratio}) of several scattering representations, on CIFAR-10 and ImageNet. For $S_C$, results are evaluated with a soft-thresholding $\rho_t$ and a thresholded rectifier $\rho_{rt}$. The last column gives
the error of  ResNet-20 for CIFAR-10 \citep{resnet} and ResNet-18 for ImageNet, taken from \url{https://pytorch.org/docs/stable/torchvision/models.html}.}
\label{scattering}
\bigskip
\makebox[\textwidth][c]{
\begin{tabular}{l l l c c c c c}
\toprule
& $\Phi$&  & $S_T$ & $S_P$ & $S_{C}\,\,(\rho_t)$  & $S_{C}\,\,(\rho_{rt})$& ResNet\\
\midrule
\multirow{2}{*}{\bf CIFAR} & Error ($\%$) & & 27.7 & 12.8 & 8.0 & 7.6 & 8.8 \\
& Fisher & & 12 & 20 & 43 & 41 & -\\
\midrule
\multirow{3}{*}{\bf ImageNet} & \multirow{2}{*}{Error ($\%$)} & Top-5 & 54.1 & 20.5 & 11.6  & 10.7 & 10.9 \\
& & Top-1 &  73.0 & 42.3 & 31.4 & 29.7 & 30.2 \\
& Fisher & & 2.0 & 18 & 51 & 44 & -\\
\bottomrule
\end{tabular}
}
\end{table}

\subsection{Separation and Concentration in Learned Scattering Networks}
\label{singl-learn}

A scattering tree iteratively separates class means with wavelet filters.
Its dimension is reduced by predefined projection operators,
which may decrease the Fisher ratio and linear separability. 
To avoid this source of inefficiency, we define a scattering network which learns these projections.
The second step introduces tight frame thresholdings along scattering channels, to concentrate within-class variabilities.
Image classification results are evaluated on the CIFAR-10 \citep{cifar} and ImageNet \citep{imagenet-dataset} datasets.

\paragraph{Learned scattering projections}
Beyond scattering trees,
the projections $P_j$ of a scattering transform
(\ref{scatree}) can be redefined as arbitrary orthogonal
$1 \times 1$ convolutional operators, which 
reduce the number of scattering channels: $P_j P_j^T = \Id$. 
Orthogonal projectors acting along the direction index $\ell$ of wavelet
filters can improve classification \citep{OyallonICLR}.
We are now going to learn these linear operators together with the
final linear classifier. Before 
computing this projection, 
the mean and variances of each scattering channel is standardized
with a batch normalization $B_N$, by setting affine coefficients $\gamma=1$ and $\beta=0$. This projected scattering operator can be written:
\[
S_P = \prod_{j=1}^J P_j\,B_N\, \rho_r\, F_w .
\]
Applying a linear classifier to $S_P(x)$ defines a deep convolutional
network whose parameters are the $1 \times 1$ convolutional $P_j$ and the classifier weights $W$, $b$. 
The wavelet convolution filters in $F_w$ are not learned. 
The orthogonality of
$P_j$ is imposed through the gradient steps (\ref{Parseval}) applied to $F = P_j^T$. Table \ref{scattering} shows that learning the projectors $P_j$
more than halves the scattering classification error of $S_P$ relatively to $S_T$ on CIFAR-10 and ImageNet, reaching AlexNet accuracy on ImageNet, while achieving a higher Fisher ratio.

The learned orthogonal projections $P_j$ create invariants to families of linear transformations along scattering channels that depend upon scales, directions and phases. They correspond to image transformations which have been linearized by the scattering transform.
Small diffeomorphisms which deform the image are examples of operators
which are linearized by a scattering transform \citep{mallatscattering}. The learned projector eliminates within-class variabilities which are not discriminative across classes. Since it is linear, it does not improve linear separability or the Fisher ratio. It takes advantage of the
non-linear separation produced by the previous scattering layers.

The operator $P_j$ is a projection on a family of orthogonal
directions which define new scattering channels, and is followed by a wavelet convolution $F_w$ along spatial variables. It defines separable convolutional filters $F_w P_j$ along space and channels. Learning $P_j$ amounts to choosing orthogonal directions so that 
$\rho_r F_w P_j$ optimizes the class means separation.
If the class distributions are invariant by rotations, the separation can
be achieved with wavelet convolutions along the direction index $\ell$
\citep{OyallonICLR}, but better results are obtained by learning these filters.
This separable scattering architecture is different from separable approximations of deep network filters in discrete cosine bases \citep{harmonic-networks} or in Fourier-Bessel bases \citep{sapiro}.
A wavelet scattering computes $\rho_r F_w P_j$ as opposed to a separable decomposition $\rho_r P_j F_w$, so the ReLU is applied in a higher dimensional space indexed by wavelet variables
produced by $F_w$. It provides explicit coordinates to analyze the mathematical properties, but it also increase the number of learned parameters as shown in \Cref{parameters}, \Cref{expdetails}.

\begin{table}[t]
\caption{Evolution of Fisher ratio across layers for the scattering concentration network $S_C$ with a ReLU with threshold $\rho_{rt}$, on the CIFAR dataset.}
\label{fisher-multiscale}
\bigskip
\makebox[\textwidth][c]{
\begin{tabular}{l l c c c c c c c c c c c c c}
\toprule
\multirow{2}{*}{\bf CIFAR} & Layer & 0 & 1 & 2 & 3 & 4 & 5 & 6 & 7 & 8\\
& Fisher & 1.8 & 11 & 13 & 11 & 15 & 15 & 22 & 25 & 40 \\
\bottomrule
\end{tabular}
}
\end{table}

\paragraph{Concentration along scattering channels}
A projected scattering transform can separate class means, but does not concentrate class variabilities. To
further reduce classification errors, following
Section \ref{Tightsec}, a concentration is computed with 
a tight frame soft-thresholding $F_j^T \rho_t F_j$, applied
on scattering channels. It increases the dimension of scattering channels with a $1 \times 1$ convolutional tight frame $F_j$,
applies a soft-thresholding $\rho_t$, and 
reduces the number of channels with the
$1 \times 1$ convolutional operator $F_j^T$. The resulting 
concentrated scattering operator is
\begin{equation}
    \label{MultiScatlearn}
S_{C} = \prod_{j=1}^J (F_j^T\,\rho_t\,F_j)\, (P_j\,B_N\, \rho_r F_w)  .
\end{equation}
It has $2J$ layers, with odd layers computed by separating means with a ReLu $\rho_r$ and even layers computed by concentrating class variabilities with a soft-thresholding $\rho_t$. According to Section \ref{Tightsec} the soft-threshold is $\lambda = 1.5 \sqrt{d/p}$.
This soft-thresholding may be replaced by a thresholded rectifier
$\rho_{rt} (u) = \max(u-\lambda,0)$ with a lower threshold $\lambda = \sqrt{d/p}$.
A logistic classifier is applied to $S_C(x)$.
The resulting deep network does not include any learned bias parameter, except in the final linear classification layer. Learning is reduced to the $1 \times 1$ convolutional operators $P_j$ and $F_j$ along scattering channels, and 
the linear classification parameters.
 
Table \ref{scattering} gives the classification errors of this concentrated
scattering on  CIFAR for $J =4$ ($8$ layers) and ImageNet for $J = 6$ ($12$ layers).
The layer dimensions are specified in Appendix \ref{expdetails}.
The number of parameters of the scattering networks are given in \Cref{parameters}, \Cref{expdetails}.
This concentration step reduces the error of $S_C$ by about $40\%$ relatively 
to a projected scattering $S_P$. A ReLU thresholding $\rho_{rt}$ produces an error slightly below a soft-thresholding $\rho_t$ both on CIFAR-10 and ImageNet, and this error is also
below the errors of ResNet-20 for CIFAR and ResNet-18 for ImageNet.
These errors are also nearly half the classification errors 
previously obtained by cascading a
scattering tree $S_T$ with several $1\times 1$ convolutional layers and
large MLP classifiers  \citep{zarka, oyallon2017scaling}. It shows that the separation and concentration learning must be done at each scale rather than at the largest scale output.
Table \ref{fisher-multiscale} shows the progressive improvement of the  Fisher ratio  measured at each layer of $S_C$ on CIFAR-10. 
The transition from an odd layer $2j-1$ to an even layer $2j$
results from $Fj^T \rho_t F_j$, which always improve the Fisher ratio by concentrating class variabilities. The transition from $2j$ to $2j+1$ is done by $P_{j+1} \rho_r F_w$, which may decrease the Fisher ratio because of the
projection $P_{j+1}$, but globally brings an important improvement.

\section{Conclusion}
We proved that separation and concentration of probability measures can be achieved with rectifiers and thresholdings applied to appropriate tight frames $F$. 
We also showed that the separation of class means can be achieved by cascading
wavelet frames that are not learned. It defines a scattering transform. 
By concentrating variabilities with a thresholding 
along scattering channels, we reach ResNet-18 classification accuracy on CIFAR-10 and ImageNet.

A major mathematical issue is to understand the mathematical properties of the learned projectors and tight frames along scattering channels. This is necessary to understand the types of classification problems that are well approximated with such architectures, and to prove lower bounds on the evolution of Fisher ratios across layers.

\subsubsection*{Acknowledgments}
This work was supported by grants from Région Ile-de-France and the PRAIRIE 3IA Institute of the French ANR-19-P3IA-0001 program. We would like to thank the Scientific Computing Core at the Flatiron Institute for the use of their computing resources. 

\bibliography{refs}
\bibliographystyle{iclr2021_conference}

\newpage
\appendix

\section{Proof of Proposition \protect \ref{prop1}}
\label{prop_proof}

We first prove the following lemma:
\begin{lemma}
  \label{fisher_linear}
  If $\Phi$ is linear, then the Fisher ratio is decreased (or equal) and the optimal linear classification error is increased (or equal).
\end{lemma}

If $\Phi$ is linear, then it is a matrix $ \in \RR^{p \times d}$. We assume that $\Phi$ has rank $p$ (and thus $p \leq d$) for the sake of simplicity. By applying a polar decomposition on $\Phi \Sigma_W^{\frac12}$, we can write
$$ \Phi = U P \Sigma_W^{-\frac12} \, , $$
where $U \in \RR^{p \times p}$ is symmetric positive-definite and $P \in \RR^{p \times d}$ verifies $P P^T = \Id$. The within-class covariance and class means of $\Phi x$ are given by
\begin{align*}
    \overline \Sigma_W &= \Phi \Sigma_W \Phi^T = U^2 \, ,\\
    \overline \mu_c &= \Phi \mu_c = U P \Sigma_W^{-\frac12} \mu_c \, .
\end{align*}
The Fisher ratio of $\Phi x$ is thus:
\begin{align*}
    C^{-1}\Tr\parenn{\overline \Sigma_W^{-1} \overline \Sigma_B} &= \Ave_c \normm{ \overline \Sigma_W^{-1/2} \bar \mu_c}^2 \\
    &= \Ave_c \normm{ P \Sigma_W^{-\frac12} \mu_c }^2 \\
    &\leq \Ave_c \normm{ \Sigma_W^{-\frac12} \mu_c}^2 \\
    &= C^{-1} \Tr\parenn{\Sigma_W^{-1} \Sigma_B} ,
\end{align*}
so $\Phi$ decreases the Fisher ratio. Besides, if $(W, b)$ is the optimal linear classifier on $\Phi x$, then $(W \Phi, b)$ is a linear classifier on $x$, and thus has a larger (or equal) error than the optimal linear classifier on $x$.

Now, if $\Phi$ has a linear inverse $\Phi^{-1}$, we apply the \Cref{fisher_linear} to $x' = \Phi x$ and $\Phi' = \Phi^{-1}$ (so that $\Phi' x' = x$), which concludes the proof.

Additionally, we can see from the proof of the lemma that a linear $\Phi$ preserves the Fisher ratio if and only if $\normm{P \Sigma_W^{-\frac12} \mu_c} = \normm{\Sigma_W^{-\frac12} \mu_c}$ for all $c$. This happens when $\Sigma_W^{-\frac12} \mu_c$ is in the orthogonal of $\Ker P = \Ker U P = \Ker \Phi \Sigma_W^{\frac12}$, which means that $\Sigma_W^{-1} \mu_c$ is in the orthogonal of $\Ker \Phi$. When $\Phi$ is an orthogonal projector, the orthogonal of $\Ker \Phi$ is the range of $\Phi$.

\section{Proof of Theorem \protect \ref{th1}}
\label{ThProof}

We begin by proving (\ref{tracW}).
Since $\Tr(\Sigma_W) = \Ave_c \Tr(\Sigma_c)$
with $\Tr(\Sigma_c)= \E(\|x_c - \mu_c \|^2)$ and 
$x_c$ is a mixture of ${\cal N}(\mu_{c,k} , \sigma^2 \Id)$ we get that
$\Tr(\Sigma_c) = \Tr(\Sigma_M) + d\, \sigma^2$ with
\[
\Tr(\Sigma_M) = C^{-1} \sum_k \pi_{c,k} \, \|\mu_{c,k} - \mu_c\|^2 ,
\]
which verifies (\ref{tracW}).

The inequalities (\ref{tracWbar}) and (\ref{meanbar})
of Theorem \ref{th1} are derived
from the following lemma which is mostly a consequence
of a theorem proved by \citet{donohojohnstone} on
soft-thresholding estimators. 

\begin{lemma}
  \label{centralLemma}
  Let $x$ be a $d$ dimensional Gaussian vector whose distribution is $\mathcal N(\mu,\sigma^2 \Id)$ with $|\mu^{(r)}| \sim r^{-s}$. 
 For all $d \geq 4$ and $\la = \sigma\,\sqrt{2 \log d}$,
  \begin{equation}
    \label{resulnsdfs}
    \E(\|\rho_t  (x) - \mu \|^2) = O(\sigma^{2-1/s} \, \log d).
  \end{equation}
\end{lemma}

Each class $x_c$ is a mixture of several $x_{c,k}$ whose distributions are
$\mathcal N (\mu_{c,k},\sigma^2 \Id)$.
We first prove the theorem by
applying this lemma to each $x_{c,k}$, and we shall then prove the lemma.

We apply (\ref{resulnsdfs}) to $x = F x_{c,k}$,
$\mu = F \mu_{c,k} = \{ \lb \mu_{c,k} , f_m \rb \}_m$, 
and $\Phi = F^T \rho F$. Since $F$ is orthogonal
\begin{equation}
  \label{ineq}
\E(\|\Phi(x_{c,k}) -  \mu_{c,k} \|^2) = 
\E(\|\rho_t F x_{c,k} - F \mu_{c,k} \|^2) = O(\sigma^{2-1/s}\, \log d) .
\end{equation}
Let $\overline \mu_{c} = \E(\Phi(x_{c}))$
  and $\overline \mu_{c,k} = \E(\Phi(x_{c,k}))$. As we have the decomposition
  \begin{equation*}
      \E(\|\Phi(x_{c,k}) -  \mu_{c,k} \|^2) = \E(\|\Phi(x_{c,k}) -  \overline \mu_{c,k} \|^2) + \|\overline \mu_{c,k} - \mu_{c,k} \|^2,
  \end{equation*}
equation (\ref{ineq}) implies that
\begin{equation}
  \label{eq1}
\|\overline \mu_{c,k} - \mu_{c,k} \|^2 = O(\sigma^{2-1/s}\, \log d)
\end{equation}
and
\begin{equation}
  \label{eq2}
\E(\|\Phi(x_{c,k}) -  \overline \mu_{c,k} \|^2) = O(\sigma^{2-1/s}\, \log d) .
\end{equation}

We first prove (\ref{meanbar}) by observing that
\[
\|\mu_c - \overline \mu_c \|^2 =
\| \sum_{k} \pi_{c,k} (\mu_{c,k} - \overline \mu_{c,k}) \|^2
\leq \Big(\sum_{k} \pi_{c,k} \|\mu_{c,k} - \overline \mu_{c,k} \| \Big)^2
\]
It results from (\ref{eq1}) that
\[
\|\mu_c - \overline \mu_c \|^2 = O(\sigma^{2-1/s}\, \log d)
\]
which proves (\ref{meanbar}).

As in the proof of (\ref{tracW}), we verify that
\[
\Tr(\overline \Sigma_W) = \Tr(\overline \Sigma_M) +
C^{-1} \sum_{c,k} \pi_{c,k}\, \E(\|\Phi(x_{c,k}) - \overline \mu_{c,k} \|^2),
\]
with
\[
\Tr(\overline \Sigma_M) = C^{-1} \sum_{c,k} \pi_{c,k} \, \|\overline \mu_{c,k} - \overline \mu_c\|^2 .
\]
Inserting (\ref{eq2}) gives
\begin{equation}
  \label{eqnisdf9}
\Tr(\overline \Sigma_W) = \Tr(\overline \Sigma_M) + O(\sigma^{2-1/s}\, \log d) .
\end{equation}
By decomposing and inserting (\ref{eq1}) we get
\begin{align*}
\Tr(\overline \Sigma_M) & \leq
C^{-1} \sum_{c,k} \pi_{c,k} \,
\Big(\|\overline \mu_{c,k} -\mu_{c,k} \| + \|\mu_{c,k} - \mu_{c} \| + \|\mu_{c} - \overline \mu_{c} \| \Big)^2 \\
& = C^{-1} \sum_{c,k} \pi_{c,k} \,
\Big(\|\mu_{c,k} -\mu_{c} \| +  O(\sigma^{1-1/(2s)}\, \log^{1/2} d)
\Big)^2\\
&= C^{-1} \sum_{c,k} \pi_{c,k} \,2\Big(
\|\mu_{c,k} -\mu_{c} \|^2 + O(\sigma^{2-1/s}\, \log d) \Big)\\
& = 2 \Tr(\Sigma_M) + O(\sigma^{2-1/s}\, \log d) .
\end{align*}

Inserting this inequality in (\ref{eqnisdf9}) proves that
\[
\Tr(\overline \Sigma_W) = 2 \Tr(\Sigma_M) + O(\sigma^{2-1/s}\, \log d)
\]
which proves (\ref{tracWbar}).

We now prove Lemma \ref{centralLemma}.
\citet{donohojohnstone} proved that
for all $d \geq 4$,
 \begin{equation}
    \label{resulnsdfs8}
    \E(\|\rho_t  (x) - \mu \|^2) \leq (2\,\log d + 1)\, (\sigma^2 +
    \sum_{m=1}^d \min (\mu[m]^2,\sigma^2) ).
 \end{equation}
We are now going to prove that if
$|\mu^{(r)}| \sim r^{-s}$ then
\[
\sum_{m=1}^d \min (\mu[m]^2,\sigma^2) = O(\sigma^{2 - 1/s}) .
\]
Let us first observe that
\begin{equation}
  \label{decay0}
\sum_{m=1}^d \min (\mu[m]^2,\sigma^2) =
\sum_{r=M+1}^d |\mu^{(r)}|^2 + M \sigma^2
\end{equation}
with $|\mu^{(M)}| \geq \sigma > |\mu^{(M+1)}|$.

Since $|\mu^{(r)}| \sim r^{-s}$, 
\[
\sum_{m=1}^d \min (\mu[m]^2,\sigma^2) \sim 
  \sum_{r=M+1}^d r^{-2s} + M \sigma^2 \sim M^{1-2s} + M \sigma^2.
\]
Since $\sigma \sim |\mu^{(M)}| \sim M^{-s}$, we conclude
\[
\sum_{m=1}^d \min (\mu[m]^2,\sigma^2) = O(\sigma^{2-1/s}) .
\]
Inserting this result in (\ref{resulnsdfs8}) 
finishes the proof of the lemma.

\section{Proof of Theorem \protect \ref{th:no-universal-approx}} \label{proof}

We choose $x = r u$ with $u \sim \mathcal U(\mathbb S^{d-1})$ and $r \in ]0, 1]$ to be determined, with $r$ and $u$ independent.
Let us fix $p \geq d$, $F \in \RR^{p \times d}$, $W \in \RR^{1 \times p}$ and $b \in \RR$. With $g(x) = W \rho_{rt} F x + b$, we have:
\begin{align*}
  g(x)
  &=   \sum_{m=1}^p w_m \rho_r\paren{r \inner{u, f_m} - \lambda} + b \\
  &=  r \sum_{m=1}^p w_m \rho_r\parenn{\inner{u, f_m} - {\lambda}/{r}} + b \, .
\end{align*}

If $\lambda = 0$, this gives $g(x) = r W \rho_r(F u) + b$ which is an affine function of $r$. Therefore, its sign can change at most once. We choose $h(x) = \cos\paren{2 \pi\norm{x}}$ so that:
\begin{equation*}
  \sgn(h(x)) =
  \begin{cases}
    +1 & r < \frac14 \text{ or } \frac34 < r \\
    -1 & \frac14 < r < \frac34
  \end{cases}
\end{equation*}
Now $g(x)$ is an affine function of $r$, so at least one of the following must occur:
\begin{equation*}
  \begin{cases}
    \sgn(g(x)) = -1 & r < \frac14 \\
    \sgn(g(x)) = +1 & \frac14 < r < \frac34 \\
    \sgn(g(x)) = -1 & \frac34 < r
  \end{cases}
\end{equation*}
We finally choose $r \sim \mathcal U(0,1)$ and so we conclude that:
$$\mathbb P\bracket{\sgn\paren{g(x)} \neq \sgn(h(x))} \geq \frac14 \, .$$

If $\lambda > 0$, then when $r \leq \lambda$, we have $\inner{u,f_m} \leq \norm{u}\norm{f_m} \leq 1 \leq \lambda/{r}$, which means that $g(x) = b$ is constant. We thus choose $r \sim \mathcal U(0, \lambda)$, $h(x) = \cos(\pi/\lambda \norm{x})$ and so we conclude that:
$$\mathbb P\bracket{\sgn\paren{g(x)} \neq \sgn(h(x))} = \frac12 \geq \frac14 \, .$$

\section{Implementation and Network Dimensions} \label{expdetails}

All networks are trained with SGD with a momentum of $0.9$ and a weight decay of ${10}^{-4}$ for the classifier weights, with no weight decay being applied to tight frames. The learning rate is set to $0.01$ for all networks, with a Parseval regularization parameter $\alpha = 0.0005$. The batch size is $128$ for all experiments. The scattering transform is based on the \emph{Kymatio} package \citep{kymatio}. Standard data augmentation was used on CIFAR and ImageNet: horizontal flips and random crops for CIFAR, and random resized crops of size $224$ and horizontal flips for ImageNet. Classification error on ImageNet validation set is computed on a single center-crop of size $224$. 

Non-linearity thresholds are set to $\lambda = 1.5\sqrt{d/p}$ for the soft-thresholding $\rho_t$, and $\lambda = \sqrt{d/p}$ for the thresholded rectifier $\rho_{rt}$. Here $d$ and $p$ represent the dimension of the patches the convolutional operators $F$ and $F^T$ act on. To ensure that the fixed threshold is well adapted to the scale of the input $x$, we normalize all its patches so that they have a norm of $\sqrt{d}$. For $1 \times 1$ convolutional operators as in $S_C$, this amounts to normalizing the channel vectors at each spatial location in $x$.

\paragraph{Two-layer networks}
When learning a frame contraction directly on the input image, $F$ is a convolutional operator over image patches of size $k\times k$ with a stride of $k/2$, where $k = 14$ for MNIST  ($d = k^2 = 196$) and $k = 8$ for CIFAR ($d= 3k^2 = 192$). The frame $F$ has $p$ output channels, where $p = 2048$ for MNIST and $p = 8192$ for CIFAR. It thus maps each patch of dimension $d$ to a channel vector of size $p \geq d$. Training lasts for $300$ epochs, the learning rate being divided by $10$ every $70$ epochs.

\paragraph{Scattering tree} We use $J = 3$ for MNIST and CIFAR and $J = 4$ for ImageNet. Each $F_w$ uses $L=8$ angles. It is followed by a standardization which sets the mean and variance of every channel to $0$ and $1$. We then learn a $1 \times 1$ convolutional orthogonal projector $P_J$ to reduce the number of channels to $d = 512$. We finally apply a $1 \times 1$ spatial normalization, as before a tight frame thresholding. Training lasts for $300$ epochs for MNIST and CIFAR ($200$ epochs for ImageNet), the learning rate being divided by $10$ every $70$ epochs ($60$ epochs for ImageNet).

\paragraph{Learned scattering}
We use $J=4$ for CIFAR and $J=6$ for ImageNet. Each $F_w$ uses $L=8$ angles. Each $P_j$ is an orthogonal projector which is a $1 \times 1$ convolution. It reduces the number of channels to $d_j$ with $d_1 = 64$, $d_2 = 128$, $d_3 = 256$ and $d_4 = 512$. For ImageNet, we also have $d_5 = d_6 = 512$. It is followed by a normalization which sets the norm across channels of each spatial position to $\sqrt{d_j}$. $F_j$ is a $1 \times 1$ convolutional tight frame with $p_j$ output channels, where  $p_1 = 1024$, $p_2 = 2048$, $p_3 = 4096$ and $p_4=8192$ for CIFAR, $p_1 = 512$, $p_2 = p_3 = 1024$ and $p_4 = p_5 = p_6 = 2048$ for ImageNet. Training lasts for $300$ epochs for CIFAR ($200$ epochs for ImageNet), the learning rate being divided by $10$ every $70$ epochs ($60$ epochs for ImageNet).

\paragraph{Fisher ratios}
Fisher ratios (\cref{Ratio}) were computed using estimations of $\Sigma_W$ and $\mu_c$ on the validation set. These estimations are unstable when the dimension $d$ becomes large with respect to the number of data samples. To mitigate this, the Fisher ratios across layers from \Cref{fisher-multiscale} were computed on the train set. Fisher ratios on ImageNet from \Cref{scattering} were computed only across channels, by considering each pixel as a distinct sample of the same class, in order to reduce dimensionality.

\begin{table}[t]
\caption{Number of parameters of scattering architectures on ImageNet. They are dominated by the size of the $1 \times 1$ orthogonal projectors $P_j$. Indeed, the wavelet tight frame $F_w$ has a redundancy of $(L + 1/4)$, whereas in ResNet strided convolutions have a redundancy of $1/2$. This is due to the fact that $F_w$ is not learned. However, $F_w$ comes with a known structure across channels, which is beneficial for the analysis of the projectors $P_j$. }
\label{parameters}
\bigskip
\makebox[\textwidth][c]{
\begin{tabular}{l l c c c c c}
\toprule
& $\Phi$& $S_T$ & $S_P$ & $S_{C}$ & ResNet-18\\
\midrule
\bf{ImageNet} & Parameters & 25.9M & 27.6M & 31.2M & 11.7M \\
\bottomrule
\end{tabular}
}
\end{table}

\end{document}